\pdfoutput=1 % MUST stay within the first 5 lines: tells arXiv to use pdfLaTeX.
\documentclass[11pt]{article}

\usepackage[font=libertinus, citestyle=numeric]{kurbanlab}

\DeclareAffiliation{hbku}{%
  College of Science and Engineering, Hamad Bin Khalifa University, Doha, Qatar}

\DeclareAffiliation{tamu}{%
  Department of Computer and Electrical Engineering,
  Texas A\&M University, College Station, TX, USA}

\DeclareAffiliation{iub}{%
  Luddy School of Informatics, Computing, and Engineering,
  Indiana University Bloomington, Bloomington, IN, USA}

\usepackage{booktabs}
\usepackage{multirow}
\usepackage[table]{xcolor}

\definecolor{posgreen}{rgb}{0.00,0.45,0.00}
\definecolor{negred}{rgb}{0.70,0.00,0.00}
\definecolor{posbg}{rgb}{0.85,0.95,0.85}
\definecolor{negbg}{rgb}{0.98,0.88,0.88}
\newcommand{\posbadge}[1]{\colorbox{posbg}{\textcolor{posgreen}{$#1$}}}
\newcommand{\negbadge}[1]{\colorbox{negbg}{\textcolor{negred}{$#1$}}}

\providecommand{\tool}[1]{\textsc{#1}}
\providecommand{\gbar}[1]{\textcolor{black!30}{\rule{#1pt}{5pt}}}
\definecolor{flipred}{RGB}{170,40,40}
\providecommand{\rev}[1]{\textcolor{flipred}{\textbf{#1}}}
\definecolor{headcol}{HTML}{1F3B63}
\definecolor{rowcol}{HTML}{EDF1F7}

\title{Accuracy and Cost Claims Do Not Survive Re-Execution in Agentic VideoQA}
\Subtitle{A significant paired conclusion reverses when the complete evaluation is run again}
\RunningTitle{Accuracy and cost claims do not survive re-execution}

\Author[equal]{Rama AlHamidi}{tamu}
\Author[equal]{Aseel Mohamed}{tamu}
\Author{Rasul Khanbayov}{hbku}
\Author{Mohamed Rayan Barhdadi}{tamu}
\Author{Erchin Serpedin}{tamu}
\Author[corresponding=hkurban@hbku.edu.qa]{Hasan Kurban}{hbku}
\ArxivID{2607.01469}        % add after you get the identifier, then re-upload

\begin{document}
\maketitle

\begin{abstract}
Agentic Video Question Answering (VideoQA) systems produce answers
through adaptive reasoning and tool-use trajectories, yet standard
practice evaluates each system once and estimates uncertainty only
across questions. This leaves a basic question untested: would the
measured method effect survive if the complete evaluation were run
again? We show that it need not. Using Static-SAGE and Dynamic-SAGE as
a controlled case study, we repeat the full paired comparison twice on
identical SAGE-Bench question--video pairs, holding the
experiment-controlled configuration, the tool library, and the scoring
protocol fixed. In the first execution, Dynamic-SAGE outperforms
Static-SAGE by $+7.33$ accuracy points; in the second, the effect
reverses to $-4.05$. Both are individually significant under standard
paired analysis, and they support opposite conclusions. The change in
the paired effect between executions is itself highly significant, and
far larger in scale than the within-execution uncertainty reported by either single execution. The reversal is consistent across question
format, modality, difficulty, and video duration, and both evaluation
arms move significantly.
Motivated by this failure, we introduce \textbf{REPAIR}
(\textbf{RE}peated \textbf{PA}ired \textbf{I}nference
\textbf{R}eliability), a protocol that repeats the paired comparison
and directly tests whether the method effect changes across complete
executions, separating \emph{directional reproducibility} from
\emph{effect-size stability}. Applied across accuracy and execution
metrics, REPAIR exposes three distinct behaviors---directional
reversal, magnitude shift, and effect attenuation---and shows that
reductions in reasoning turns and visible tool calls do not imply
reproducible reductions in primitive computation or latency. The
execution-level movement we observe is comparable to, and in this case
larger than, the median gain reported by recent agentic VideoQA
systems, which contextualizes its magnitude without implying that
those systems are themselves unstable. Significance within a single
agentic execution is not sufficient evidence that a reported method
effect is reproducible.
\end{abstract}

\printkeywords

%%%%%%%%%%%%%%%%%%%%%%%%%%%%%%%%%%%%%%%%%%%%%%%%%%%%%%%%%%%%%%%%%%%%%%%%%%%%%%%%
%% BODY
%%%%%%%%%%%%%%%%%%%%%%%%%%%%%%%%%%%%%%%%%%%%%%%%%%%%%%%%%%%%%%%%%%%%%%%%%%%%%%%%
\begin{figure}[tp]
    \centering
    % Place the Figure-1 PDF at figures/teaser.pdf (falls back to a
    % placeholder box if absent). teaser_v4.png is the older raster version.
    \kilgraphics[0.48\linewidth]{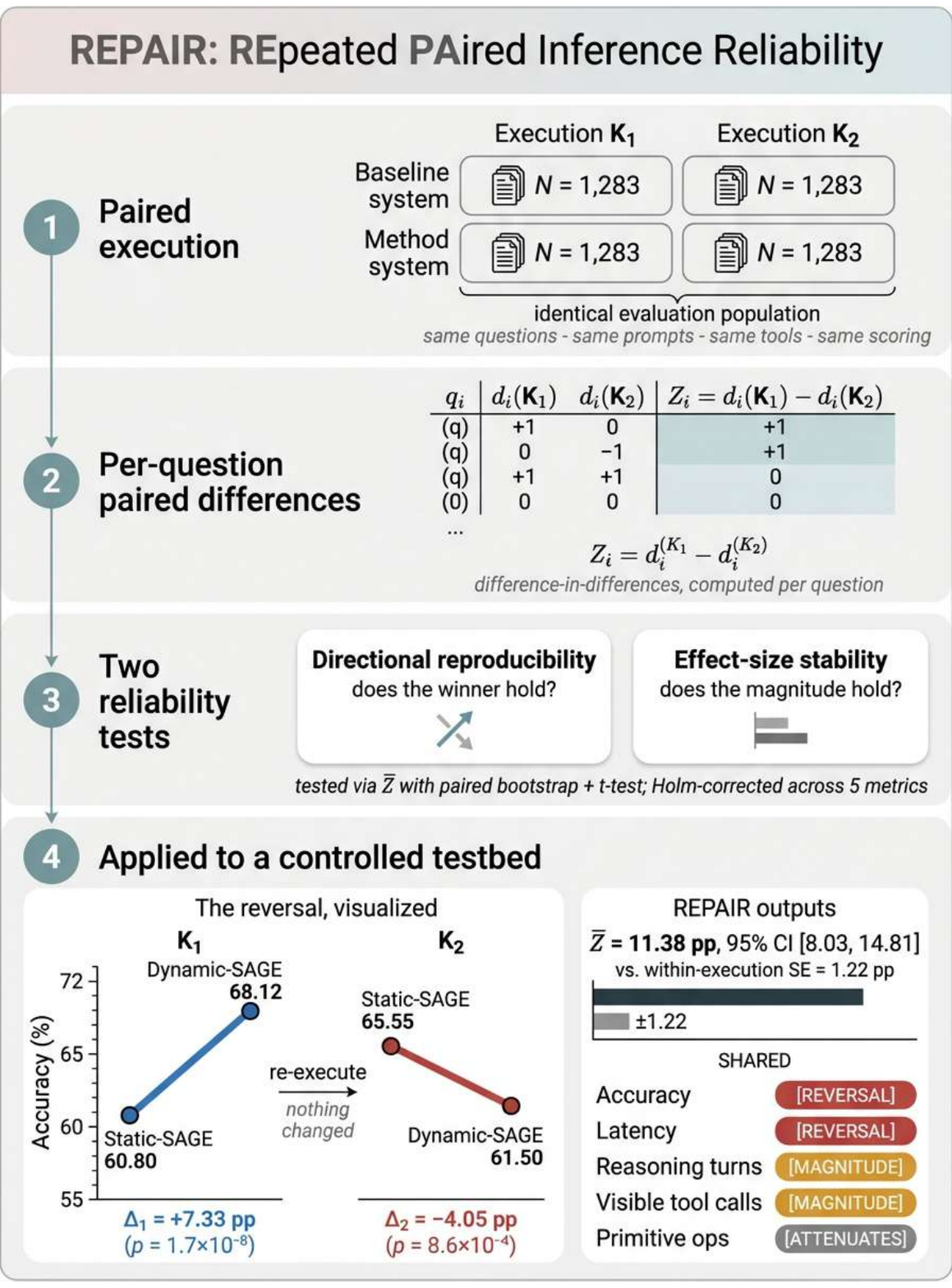}
    \caption{\textbf{REPAIR: repeated paired evaluation of an agentic
    method claim.}
    \textbf{(1)}~REPAIR evaluates a baseline and a method across two
    complete executions $K_1$ and $K_2$ on an identical population of
    $N{=}1{,}283$ SAGE-Bench question--video pairs, holding prompts,
    tool libraries, and scoring fixed.
    \textbf{(2)}~Because the same questions appear in every
    system--execution cell, the change in the paired effect is measured
    per question as
    $Z_i = d_i^{(K_1)} - d_i^{(K_2)}$, yielding a testable
    difference-in-differences rather than a comparison of aggregates.
    \textbf{(3)}~REPAIR separates \emph{directional reproducibility} from \emph{effect-size stability}.
    \textbf{(4)}~Applied to Dynamic-SAGE against Static-SAGE, a
    controlled pair differing only in a fixed library of composite
    tools, a significant $+7.33$\,pp advantage in $K_1$ becomes a
    significant $-4.05$\,pp disadvantage in $K_2$. The cross-execution
    interaction is $\bar Z = 11.38$\,pp,
    roughly nine times the $1.22$\,pp within-execution standard error
    that a single run reports. Across five metrics REPAIR returns three
    distinct verdicts---reversal, magnitude shift, and attenuation.}
    \label{fig:teaser}
\end{figure}

\section{Introduction}
\label{sec:introduction}

Agentic Video Question Answering (VideoQA) systems increasingly solve
long-video problems through iterative reasoning, evidence selection,
and tool use rather than a single forward pass~\cite{videoagent,sage}.
This adaptivity is useful because answer-bearing evidence is often
sparse and question-dependent: a brief visual event, spoken phrase, or
short temporal segment may determine the answer while most of the
video is irrelevant. Agentic systems can search for such evidence
selectively, but they also make evaluation inherently execution
dependent. Running the same configured agent again can produce a new
reasoning trajectory, new tool selections, new intermediate
observations, and potentially a different final answer.

Current evaluation practice does not generally treat the complete
execution as a source of uncertainty. A method and baseline are
typically run once on the same benchmark questions, after which paired
confidence intervals or significance tests quantify variation across
those questions. Pairing is valuable; it controls for question-level
difficulty and can establish whether two systems differ within the
observed run. But it leaves a separate question unanswered:
\emph{would the estimated method effect itself survive if the complete
agentic evaluation were executed again?}

We show that this distinction can reverse the scientific conclusion.
We audit Static-SAGE and Dynamic-SAGE on the same 1,283 SAGE-Bench
question--video pairs. The two systems share the same orchestrator and
primitive capabilities; Dynamic-SAGE additionally exposes five fixed,
previously synthesized composite tools. We repeat the complete paired
comparison twice while holding the experiment-controlled
configuration fixed. In execution $K_1$, Dynamic-SAGE appears to
provide a strong improvement over Static-SAGE, with an accuracy effect
of $+7.33$ percentage points. In $K_2$, the same comparison reverses
to $-4.05$ points. Both effects are individually significant under the
standard paired analysis. Figure~\ref{fig:teaser} summarizes this reversal and the corresponding
cross-execution comparison.

The reversal is not explained by ordinary question-level uncertainty.
Because the same questions appear in every system--execution
condition, we can test the change in the paired method effect directly.
The resulting cross-execution interaction is $11.38$ percentage
points, with 95\% CI $[+8.03,+14.81]$ and
$p=6.11\times10^{-11}$. By contrast, the mean within-execution
standard error is only $1.22$ points. Thus, the paired test is not
incorrect; it answers a narrower question. It measures uncertainty
over questions conditional on one realized execution, while the
scientific claim is often interpreted as though the estimated method
effect were stable beyond that realization.

Several experimental confounds can also be ruled out. The evaluation
population, synthesized tool library, and scoring procedure are fixed
across executions, so the shift cannot be attributed to changed
questions, regenerated tools, or a changed scoring procedure. Nor is it
confined to one evaluation arm or subgroup: Static-SAGE and
Dynamic-SAGE both change significantly across executions, and the
Dynamic-minus-Static effect changes from positive to negative across
all 11 examined subgroups spanning question format, modality,
difficulty, and video duration.

Dynamic-SAGE also provides a particularly controlled setting in which
to expose this failure. Static-SAGE and Dynamic-SAGE share the same
orchestrator, inference loop, and primitive capabilities; the
intervention is the addition of a fixed, offline-synthesized composite
tool library. Re-evaluating this comparison therefore lets us audit not
only whether the originally observed Dynamic-SAGE advantage reproduces,
but whether a method effect associated with a well-defined change in
the agent's action space remains stable under complete re-execution.

Motivated by this failure, we introduce
\textbf{REPAIR} (\textbf{RE}peated \textbf{PA}ired
\textbf{I}nference \textbf{R}eliability), a protocol for evaluating
whether agentic method claims survive complete re-execution. REPAIR
retains question-level pairing but repeats the full paired comparison,
making the execution-specific method effect itself an object of
evaluation. It directly tests the cross-execution interaction and
separates two properties that a single run conflates:
\emph{directional reproducibility}, whether the supported direction of
a claim persists, and \emph{effect-size stability}, whether its
magnitude remains consistent.

Applying REPAIR beyond accuracy reveals that execution sensitivity is
not uniform. Accuracy and latency reverse direction across executions;
reductions in reasoning turns and visible tool calls reproduce
directionally but change significantly in magnitude; and the increase
in primitive operations observed in $K_1$ attenuates to no measurable
effect in $K_2$. This analysis also exposes a second evaluation issue:
visible tool calls and reasoning turns characterize orchestration, not
necessarily consumed resources. Dynamic-SAGE reduces visible calls in
both executions, yet this does not translate into reproducible
reductions in primitive computation or runtime.

The observed execution-scale movement is also comparable to gains
reported by recent agentic VideoQA systems, motivating explicit
treatment of execution as an evaluation dimension.

Our contributions are:

\begin{itemize}
    \item \textbf{A direct demonstration that single-execution
    significance need not imply a reproducible agentic method effect.}
    On the same 1,283 SAGE-Bench questions, a significant
    $+7.33$-point Dynamic-SAGE advantage becomes a significant
    $-4.05$-point disadvantage under re-execution. The resulting
    $11.38$-point cross-execution interaction is itself highly
    significant.

    \item \textbf{REPAIR and a controlled audit of Dynamic-SAGE.} We introduce REPAIR, a repeated paired evaluation protocol, and instantiate it through a controlled re-evaluation of Dynamic-SAGE against Static-SAGE. Because the comparison preserves the orchestrator and primitive capabilities while changing the available composite-tool action space, it provides a clean test of whether an observed method effect survives complete re-execution. REPAIR separates directional reproducibility from effect-size stability.

    \item \textbf{A multi-metric audit of execution sensitivity in
    agentic VideoQA.}
    We show that re-execution produces distinct behaviors---directional
    reversal, magnitude sensitivity, and effect attenuation---and that
    reductions in orchestration steps do not necessarily correspond to
    reproducible reductions in primitive computation or runtime.
\end{itemize}

\section{Related Work}
\label{sec:related_work}

% =====================================================================
\subsection{Agentic Video Question Answering}

VideoQA has progressed toward long-context, temporally grounded, and
multimodal reasoning, where answer-bearing evidence may occupy only a
small portion of a long video~\cite{sage,videoagent}. Modular methods
such as MoReVQA~\cite{morevqa} and ProViQ~\cite{proviq} decompose
VideoQA into intermediate operations over selected evidence. Recent systems increasingly use iterative agents, specialized tools,
multi-agent collaboration, and adaptive evidence acquisition for
long-video reasoning
\cite{videoagent,lvagent,starvideoquestionanswering,vital,longvideoagent,symphony,videoseal}.
SAGE~\cite{sage}, which forms the basis of our audit, uses an
any-horizon agent that may answer directly or escalate to iterative
tool use. Although these systems differ architecturally, their outputs
emerge from realized reasoning and evidence-acquisition trajectories;
our work concerns the reliability of this shared evaluation layer.

% =====================================================================
\subsection{Tool Use and Reusable Program Synthesis}

Agentic VideoQA builds on modular and program-generating vision
systems. VisProg~\cite{VisProg} composes predefined visual modules,
ViperGPT~\cite{vipergpt} synthesizes executable Python programs, and
VADAR~\cite{vadar} constructs callable procedures for visual spatial
reasoning. Related long-video approaches such as META~\cite{meta}
likewise convert recurring tool trajectories into reusable higher-level
operations. Dynamic-SAGE, our audit target, uses a persistent offline variant of
this idea: a fixed library of composite tools is synthesized from
existing SAGE primitives before evaluation and reused unchanged across
unseen questions. We do not claim synthesis novelty or superiority;
the Static-versus-Dynamic comparison serves only as a controlled
intervention for studying whether an observed method effect survives
re-execution.

% =====================================================================
\subsection{Statistical Evaluation and Reproducibility}

Repeated evaluation is increasingly recognized as important for
stochastic language-model and agentic systems. Prior work quantifies
variation across repeated LLM benchmark evaluations
\cite{blackwell2024reproducible}, repeated interactive-agent trials
\cite{taubench}, and repeated SWE-Bench trajectories
\cite{randomness_agentic_evals}, showing that single-run scores can
mask meaningful run-to-run variation. Agentic VideoQA has begun to recognize the same issue. SVAgent
\cite{svagent} reports ten independently seeded evaluations on MLVU,
including repeated-run means and standard deviations, providing a
direct precedent for evaluating an agentic VideoQA system over more
than one execution. Such repeated-score reporting addresses whether a
system's absolute performance is stable across runs. REPAIR targets a complementary inferential question: whether the
\emph{paired method effect} is stable. Rather than considering
baseline and method scores separately, we preserve the same questions
across both systems and executions and analyze
$\hat{\Delta}_k$, the paired method effect within execution $K_k$.
The cross-execution interaction then tests whether
$\hat{\Delta}_k-\hat{\Delta}_{\ell}$ differs from zero. This
distinguishes variability in absolute system performance from
variability in the scientific comparison used to support a method
claim. To our knowledge, prior agentic VideoQA work has not
operationalized this question-paired cross-execution interaction for
the method effect.
\begin{figure}[tp]
    \centering
    \kilgraphics[\linewidth]{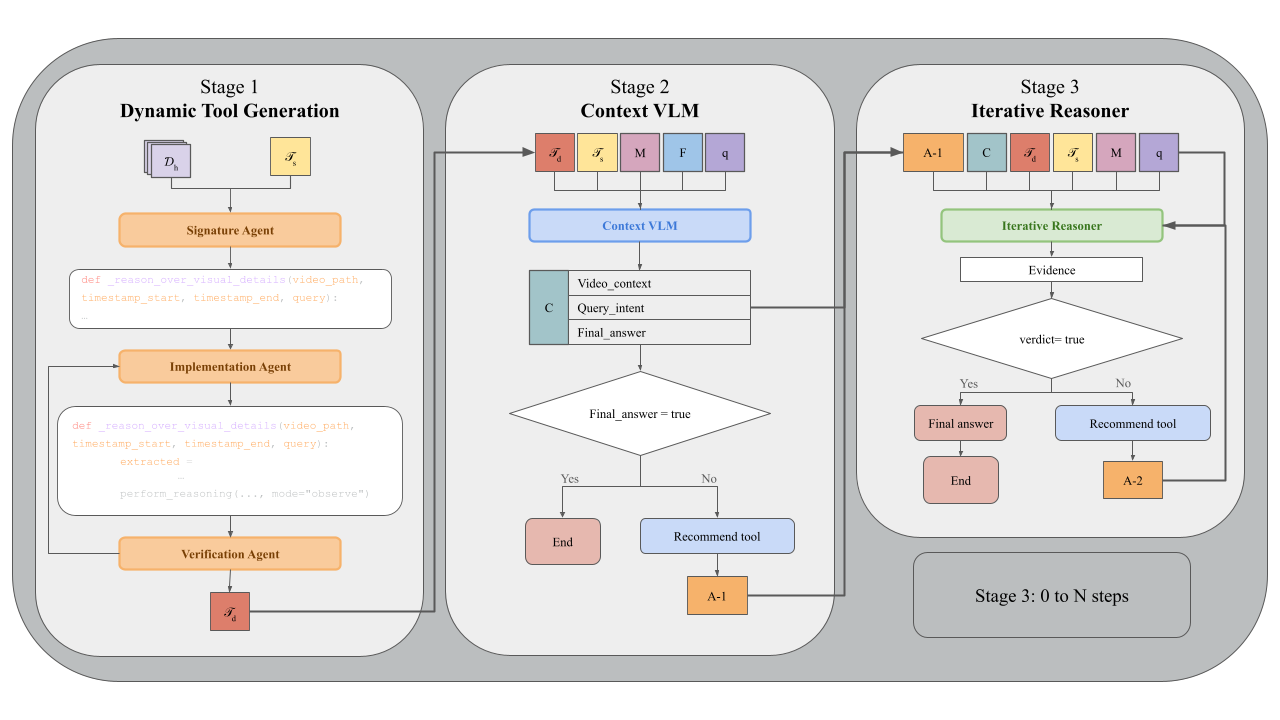}
    \caption{\textbf{Overview of Dynamic-SAGE.} Dynamic-SAGE comprises three stages: (1) offline dynamic tool synthesis, (2) Context VLM, and (3) Iterative Reasoner. The offline synthesis stage takes held-out queries $\mathcal{D}_{h}$ (excluded from evaluation) and the static tool set $\mathcal{T}_{s}$ as input. The \textit{Signature Agent} proposes tool signatures, which the \textit{Implementation Agent} turns into function bodies by composing static tools. The \textit{Verification Agent} then validates each implementation for safety and contract compliance, returning failures to the Implementation Agent for retry; surviving tools are fingerprinted and deduplicated to form the dynamic tool library $\mathcal{T}_{d}$. Stages 2 and 3 reuse SAGE's orchestration loop but operate over the augmented library: the \textit{Context VLM} receives the dynamic and static tools, video metadata (M), up to 64 sampled frames (F), and the query (q) to build video context and decide, based on the query's reasoning horizon, whether to answer directly or invoke iterative tool use. When the \textit{Iterative Reasoner} is invoked, context and media from previous steps are injected; tools return observations as evidence, and only the orchestrator commits the final answer.}
    \label{fig:overview}
\end{figure}

\section{Methods}
\label{sec:methods}

We study the reliability of paired comparisons between agentic
VideoQA systems under complete re-execution. Our method has two
components. First, we formalize the distinction between uncertainty
over benchmark questions within one execution and variation of the
estimated method effect across executions. Second, we introduce
REPAIR, which repeats the complete paired comparison and directly
tests whether the method effect itself changes. We instantiate this
protocol on Static-SAGE and Dynamic-SAGE, which provide a controlled
comparison between identical primitive capabilities with and without
a fixed library of synthesized composite tools.

% =====================================================================
\subsection{Paired Evaluation of Agentic Systems}
\label{sec:paired_agentic_eval}

Let a VideoQA evaluation set be
$
    \mathcal{D}
    =
    \{(V_i,q_i,a_i)\}_{i=1}^{N},
$ where $V_i$ is a video, $q_i$ is a question, and $a_i$ is the
ground-truth answer. Some questions additionally provide a candidate
answer set $\mathcal{C}_i$, while open-ended questions use
$\mathcal{C}_i=\varnothing$. Unlike a single-pass predictor, an agentic VideoQA system produces its
answer through a realized sequence of reasoning decisions and tool
calls. Let $\mathcal{O}$ denote the orchestrator and $\mathcal{T}$ its
registered tool library. A complete execution produces
$
    \hat a_i
    =
    \mathcal{O}
    (V_i,q_i,\mathcal{C}_i;\mathcal{T}),
$ together with the corresponding reasoning and tool-use trace. We consider a paired comparison between a baseline system $s$ and a
modified system $d$, evaluated on the same $N$ question--video pairs.
For scalar outcome $m$, let
$x^{s,(m)}_{ik}$ and $x^{d,(m)}_{ik}$ denote the outcome for question
$i$ in complete execution $K_k$. Their question-level paired
difference is
$
    d^{(m)}_{ik}
    =
    x^{d,(m)}_{ik}
    -
    x^{s,(m)}_{ik},
$
and the method effect estimated within that execution is
$
    \hat{\Delta}^{(m)}_k
    =
    \frac{1}{N}
    \sum_{i=1}^{N}
    d^{(m)}_{ik}.
    \label{eq:execution_effect}
$
For accuracy,
$x^{s,(\mathrm{acc})}_{ik},
x^{d,(\mathrm{acc})}_{ik}\in\{0,1\}$, so
$\hat{\Delta}^{(\mathrm{acc})}_k$ is the paired difference in
accuracy.
Standard paired inference estimates uncertainty in
$\hat{\Delta}^{(m)}_k$ from variation across the $N$ questions within
one realized execution. This answers whether the two systems differ
\emph{conditional on that execution}. It does not establish whether
the same method effect would be obtained if the complete agentic
evaluation were executed again. In an agentic pipeline, a new
execution may realize different reasoning trajectories, tool
selections, intermediate observations, and final answers even when the
experiment-controlled configuration is unchanged.

% =====================================================================
\subsection{REPAIR: Repeated Paired Evaluation}
\label{sec:repair}

We introduce \textbf{REPAIR} (\textbf{RE}peated \textbf{PA}ired
\textbf{I}nference \textbf{R}eliability), which adds execution as an
explicit evaluation dimension while preserving question-level
pairing. Rather than reporting one paired effect, REPAIR repeats the
complete comparison for $K\geq2$ executions and retains the
execution-specific effects
$
    \left\{
        \hat{\Delta}^{(m)}_1,
        \ldots,
        \hat{\Delta}^{(m)}_K
    \right\}.
$
REPAIR is an evaluation protocol rather than a new statistical
estimator. Its key design choice is to preserve question-level pairing
across complete executions and make the execution-specific method
effect, rather than the absolute system scores, the repeated object of
inference.

\paragraph{Cross-execution interaction.}
For executions $K_k$ and $K_\ell$, REPAIR measures whether the paired
method effect changes by defining
$
    Z^{(m;k,\ell)}_i
    =
    d^{(m)}_{ik}
    -
    d^{(m)}_{i\ell}.
    \label{eq:repair_interaction}
$
Averaging over the same questions gives
$
    \bar Z^{(m;k,\ell)}
    =
    \hat{\Delta}^{(m)}_k
    -
    \hat{\Delta}^{(m)}_\ell.
    \label{eq:repair_interaction_mean}
$
The usual paired analysis asks whether
$\hat{\Delta}^{(m)}_k=0$ within one execution. REPAIR additionally
asks whether
$\bar Z^{(m;k,\ell)}=0$: whether the estimated method effect itself
changes when the complete evaluation is repeated. Accordingly, inference on $\bar Z$ tests whether the two observed
execution-specific effects differ over the question population; it
does not provide a prediction interval for future executions or
estimate their population distribution. The statistical tests used for our $K=2$ instantiation are specified
in Section~\ref{sec:statistical_analysis}.

\paragraph{Execution-level spread.}
To describe the scale of the observed execution-to-execution movement,
REPAIR reports
$
    s_K^{(m)}
    =
    \sqrt{
        \frac{
            \sum_{k=1}^{K}
            \left(
                \hat{\Delta}^{(m)}_k
                -
                \bar{\Delta}^{(m)}
            \right)^2
        }{K-1}
    },
    \label{eq:repair_spread}
$
where
$\bar{\Delta}^{(m)}
=
K^{-1}\sum_{k=1}^{K}\hat{\Delta}^{(m)}_k$.
For two executions,
$
    s_K^{(m)}
    =
    \frac{
        \left|
            \hat{\Delta}^{(m)}_1
            -
            \hat{\Delta}^{(m)}_2
        \right|
    }{\sqrt{2}}.
$
With $K=2$, this quantity has only one between-execution degree of
freedom and is therefore descriptive rather than a precise estimate
of population execution variance.

\paragraph{Interpreting repeated effects.}
REPAIR separates reproducibility of a claim's \emph{direction} from
stability of its \emph{magnitude}. A claim exhibits
\textbf{directional reversal} when supported effects occur in opposite
directions across executions. It exhibits
\textbf{directional reproduction with magnitude sensitivity} when the
same direction persists but the cross-execution interaction is
nonzero. We use \textbf{effect attenuation} when an effect supported
in one execution becomes indistinguishable from zero in another.
When no cross-execution shift is detected, we do not interpret this as
proof of equivalence. These distinctions matter particularly for agentic resource claims.
Reasoning turns and visible tool calls describe the structure of an
orchestration trajectory, whereas primitive operations and wall-clock
latency more directly reflect work performed. REPAIR therefore treats
these measurements separately rather than assuming that a reduction
in orchestration steps implies a reduction in consumed resources.

\subsection{Controlled Audit Target: Static-SAGE and Dynamic-SAGE}
\label{sec:audit_target}
We instantiate REPAIR on the comparison between Static-SAGE and
Dynamic-SAGE. This pair provides a controlled agentic intervention:
the two systems use the same reasoning orchestrator and underlying
primitive capabilities, while Dynamic-SAGE additionally exposes a
fixed library of synthesized composite tools. Figure~\ref{fig:overview} summarizes the Dynamic-SAGE
architecture used as our audit target, including its offline tool-synthesis
stage and the shared SAGE inference loop. Let $\mathcal{T}_s$ denote the original Static-SAGE primitive tool
library and let $\mathcal{D}_h\subset\mathcal{D}$ denote a held-out
synthesis set excluded from evaluation. Before evaluation, an offline
synthesis procedure $\mathcal{S}$ constructs
$
    \mathcal{T}_d
    =
    \mathcal{S}
    (\mathcal{D}_h,\mathcal{T}_s),
$
where every $g\in\mathcal{T}_d$ is composed exclusively from
operations already available in $\mathcal{T}_s$. Dynamic-SAGE learns
no new model weights and introduces no new primitive capability. Using the same orchestrator $\mathcal{O}$, the two systems therefore
differ only in their registered action spaces: $  \hat a_i^{\,s}
    \mathcal{O}
    (V_i,q_i,\mathcal{C}_i;\mathcal{T}_s),
    \hat a_i^{\,d}
    \mathcal{O}
    (V_i,q_i,\mathcal{C}_i;
    \mathcal{T}_s\cup\mathcal{T}_d).
$
Dynamic-SAGE constructs $\mathcal{T}_d$ offline using the original
three-stage synthesis pipeline. A \emph{Signature Agent} proposes
reusable interfaces from recurring procedures in the held-out
questions; an \emph{Implementation Agent} realizes each interface as a
composition of existing SAGE primitives; and a
\emph{Verification Agent} rejects malformed, invalid, unsafe,
verdict-committing, or duplicate candidates before registration.
Full synthesis prompts, validation rules, and filtering statistics are
provided in the supplementary material. The accepted tools are persistent. They are registered before
evaluation and remain unchanged for all unseen questions; no tool is
generated or modified for an individual evaluation sample.
Dynamic-SAGE is therefore \emph{offline-dynamic}: it expands the
action space before inference rather than synthesizing programs
per query. At inference time, both systems use the same multi-turn SAGE reasoning
loop. Static-SAGE selects from $\mathcal{T}_s$, whereas Dynamic-SAGE
selects from $\mathcal{T}_s\cup\mathcal{T}_d$. A primitive invocation
behaves identically in either system. When Dynamic-SAGE selects a
composite tool, the orchestrator observes one tool call while the tool
internally executes its fixed composition of primitive operations and
returns evidence through the same tool-result interface. Thus,
composite tools alter the \emph{granularity of the available action
space}, which is why our evaluation distinguishes visible tool calls
from the primitive operations executed underneath them. 
\section{Experimental Setup}
\label{sec:experiments}

% =====================================================================
\subsection{Benchmark and Re-Execution Protocol}
\label{sec:benchmark_protocol}

We instantiate REPAIR on \textbf{SAGE-Bench}~\cite{sage}, auditing
the paired comparison between Static-SAGE and Dynamic-SAGE.
SAGE-Bench contains 1,744 VideoQA questions spanning seven duration
buckets (0--60\,s to 2400+\,s), three modalities (visual, verbal,
both), four difficulty levels (easy, medium, hard, very hard), and
two question formats (MCQ and open-ended). The original Dynamic-SAGE pipeline reserves a stratified
$H{=}150$-question synthesis holdout $\mathcal{D}_h$, which is
excluded from evaluation. After removing this holdout and samples whose
externally hosted source videos were unavailable during the original
evaluation, 1,304 questions remained. We execute the complete Static-versus-Dynamic comparison twice,
denoting the two executions by $K_1$ and $K_2$. At re-execution time,
21 additional source videos were unavailable. To ensure that execution
effects cannot be explained by changes in the evaluation population,
all cross-execution analyses use the exact intersection of
\textbf{1,283 question--video pairs} observed in both executions.
Static-SAGE and Dynamic-SAGE are therefore compared on the same
questions within each execution, and $K_1$ and $K_2$ are compared on
the same evaluation population.
\begin{table}[t]
\centering
\footnotesize
\setlength{\tabcolsep}{4pt}
\renewcommand{\arraystretch}{1.15}
\caption{\textbf{Primitive tool library $\mathcal{T}_s$.}
Each entry is a typed, single-step operation with no internal planning.
Dynamic-SAGE inherits $\mathcal{T}_s$ unchanged and introduces only
\emph{verified compositions} of these primitives, so no new capability
enters the system at composition time.}
\label{tab:tools}
\begin{tabular}{@{}ll>{\raggedright\arraybackslash}p{2.75cm}@{}}
\toprule
\textbf{Tool} & \textbf{Signature} & \textbf{Primitive operation} \\
\midrule
\multicolumn{3}{@{}l}{\emph{Evidence acquisition}} \\
\tool{WebSearch}         & query $\to$ docs            & Retrieve web evidence for a query \\
\tool{ParseWebsite}      & url $\to$ text              & Extract readable page content \\
\tool{TranscribeSpeech}  & audio $\to$ text$_{\tau}$   & Time-aligned speech transcript \\
\addlinespace[3pt]
\multicolumn{3}{@{}l}{\emph{Video grounding and reasoning}} \\
\tool{GroundEvent}       & video, query $\to \tau$     & Localize an event to a time span \\
\tool{ExtractVideoParts} & video, $\tau \to$ frames    & Sample frames or cut a clip \\
\tool{Analyze}           & frames, text $\to$ answer   & Joint reasoning over media and text \\
\bottomrule
\end{tabular}
\end{table}

% =====================================================================
\subsection{Systems and Controlled Configuration}
\label{sec:systems_configuration}

Both conditions use \texttt{GPT-4o}~\cite{GPT4o} as the reasoning
orchestrator and the same multi-turn SAGE inference loop. Static-SAGE
selects from the six primitive tools in $\mathcal{T}_s$
(Table~\ref{tab:tools}), while Dynamic-SAGE selects from
$\mathcal{T}_s\cup\mathcal{T}_d$. The synthesized library
$\mathcal{T}_d$ contains five fixed composite tools
(Table~\ref{tab:synth_tools}), each constructed only from operations
already available in $\mathcal{T}_s$. The same synthesized library is reused in $K_1$ and $K_2$ without
regeneration. Across the two executions, we hold fixed all
experiment-controlled configuration: system and orchestration prompts,
decoding configuration, iterative reasoning budget, primitive and
synthesized tool libraries, tool-registration procedure, and scoring
protocol. Both executions use the same nominal \texttt{GPT-4o}
configuration. The runs are nevertheless separate end-to-end executions. Reasoning
trajectories, tool selections, tool invocations, intermediate
observations, and final responses are generated anew rather than
replayed from cached outputs. We do not assume that unobservable
provider-side serving state or mutable external-tool state is identical
across executions; our target is reproducibility of the complete
realized evaluation under a fixed experiment-controlled configuration.
Decoding temperatures and reasoning budgets are reported in
Section~S1 of the supplementary material.

\begin{table}[t]
\centering
\footnotesize
\setlength{\tabcolsep}{4pt}
\renewcommand{\arraystretch}{1.15}
\caption{\textbf{Fixed synthesized library $\mathcal{T}_d$.}
Each tool is a frozen composition of the primitives in Tab.~\ref{tab:tools},
abbreviated here by their leading verb. Notation: $\rightarrow$ is a data
dependency, $\{\cdot\}$ groups operations executed at the same stage, and
$d_k$ counts primitive calls. No entry introduces a capability absent
from $\mathcal{T}_s$.}
\label{tab:synth_tools}
\begin{tabular}{@{}llc@{}}
\toprule
\textbf{Tool} & \textbf{Composition} & $\boldsymbol{d_k}$ \\
\midrule
\multicolumn{3}{@{}l}{\emph{Evidence pairing (no reasoning stage)}} \\
\tool{VisualVerbal} & $\{$\tool{Extract}, \tool{Transcribe}$\}$        & 2 \\
\tool{TopicSpeech}  & \tool{Ground} $\rightarrow$ \tool{Transcribe}    & 2 \\
\addlinespace[3pt]
\multicolumn{3}{@{}l}{\emph{Reasoning-terminated}} \\
\tool{VisualDetail} & \tool{Extract} $\rightarrow$ \tool{Analyze}      & 2 \\
\tool{McqEvidence}  & $\{$\tool{Extract}, \tool{Transcribe}$\}
                       \rightarrow$ \tool{Analyze}                     & 3 \\
\tool{EventDetail}  & \tool{Ground} $\rightarrow$ \tool{Extract}
                       $\rightarrow$ \tool{Analyze}                    & 3 \\
\midrule
\multicolumn{2}{@{}l}{Mean composition depth $\bar{d}$} & 2.4 \\
\bottomrule
\end{tabular}
\end{table}

% =====================================================================
\subsection{Measurements and Statistical Instantiation}
\label{sec:statistical_analysis}

For every question, system, and execution, we record correctness,
reasoning turns, visible tool calls, primitive operations, and
end-to-end latency. Visible calls count orchestrator-issued
invocations, whereas primitive operations count the underlying
functions executed; we therefore report them separately. Correctness is scored by a temperature-0 GPT-4o semantic judge applied
once to the pooled outputs of both executions, ensuring that identical
responses receive identical labels. Judge-family robustness is
reported in the supplementary material. We instantiate REPAIR with $K=2$ executions and $N=1{,}283$
matched questions. Execution-specific effects use paired-bootstrap
95\% confidence intervals and, for accuracy, McNemar's test.
Cross-execution interactions use 10,000 paired bootstrap resamples and
paired $t$-tests on the question-level interaction values
$Z_i^{(m)}$. Because multiple questions may originate from the same source video,
we additionally repeat the accuracy analyses with a video-clustered
bootstrap. The 1,283 questions correspond to 922 unique source videos.
For each of 10,000 resamples, we sample 922 videos with replacement
and retain all questions associated with each sampled video, preserving
the four-way system--execution pairing. The central cross-execution interaction remains nonzero under this
analysis (95\% CI $[+8.08,+14.85]$), as do the execution-specific
and within-method accuracy effects; full video-clustered results are
reported in Section~S4 of the supplementary material. We test five primary cross-execution outcomes---accuracy, reasoning
turns, visible tool calls, primitive operations, and latency---and
control family-wise error using Holm's procedure at $\alpha=0.05$.
Holm correction does not require independence among these tests, which
is useful because the execution metrics arise from the same
trajectories. All five interactions remain significant after
correction. For accuracy we additionally report the descriptive spread
$s_K^{(\mathrm{acc})}$; because $K=2$ provides only one
between-execution degree of freedom, we do not interpret it as a
precise population variance estimate.

\section{Results}
\label{sec:results}

% =====================================================================

\begin{table}[t]
\centering
\footnotesize
\setlength{\tabcolsep}{4.5pt}
\renewcommand{\arraystretch}{1.15}
\caption{\textbf{The Dynamic-SAGE advantage reverses sign under
re-execution.} $K_1$ and $K_2$ are two executions over the identical
1{,}283 question--video pairs, with the same tool libraries, prompts, and
backbone; only the execution differs. $\Delta$ is Dynamic minus Static
accuracy in percentage points (pp), with paired bootstrap intervals and
exact McNemar $p$-values over item-level correctness. Both effects are
individually significant and the two intervals are disjoint, so the
reversal cannot be attributed to sampling noise in either run. Bold marks
the stronger system per execution.}
\label{tab:reexecution_accuracy}
\begin{tabular*}{\linewidth}{@{\extracolsep{\fill}}lrrrcc@{}}
\toprule
& \multicolumn{2}{c}{\textbf{Accuracy (\%)}}
& \multicolumn{3}{c}{\textbf{Dynamic $-$ Static}} \\
\cmidrule(lr){2-3} \cmidrule(l){4-6}
\textbf{Exec.} & \textbf{Static} & \textbf{Dynamic}
& $\boldsymbol{\Delta}$ \textbf{(pp)} & \textbf{95\% CI} & $\boldsymbol{p}$ \\
\midrule
$K_1$ & 60.80 & \textbf{68.12} & $+7.33$ & $[+4.83,+9.90]$ & $1.7{\times}10^{-8}$ \\
$K_2$ & \textbf{65.55} & 61.50   & $-4.05$ & $[-6.39,-1.79]$ & $8.6{\times}10^{-4}$ \\
\midrule
\multicolumn{3}{@{}l}{Swing $\Delta(K_1)-\Delta(K_2)$}
& $11.38$ & \multicolumn{2}{c}{---} \\
\bottomrule
\end{tabular*}
\end{table}

\begin{table}[t]
\centering
\footnotesize
\setlength{\tabcolsep}{4pt}
\renewcommand{\arraystretch}{1.12}
\caption{\textbf{Re-execution variation in our system is comparable to, or
larger than, the headline gains reported by recent agentic VideoQA
methods.} Upper block: absolute within-paper accuracy improvements from
matched baseline--method comparisons under the same benchmark and
evaluation setting, used only to calibrate what counts as a large effect
in this literature. These are \emph{not} estimates of re-execution
variability for those systems. Lower block reports the observed movement of the paired
Dynamic-SAGE--Static-SAGE method effect across two complete
executions of the same experiment-controlled comparison (Tab.~\ref{tab:reexecution_accuracy}). The comparison is therefore between a typical reported improvement and
the amount by which a paired method effect can move under complete
re-execution. Full provenance for each
comparison is given in the supplementary material.}
\label{tab:recent_agentic_gains}
\begin{tabular*}{\linewidth}{@{\extracolsep{\fill}}lr@{\hspace{6pt}}l@{}}
\toprule
\textbf{System} & \textbf{Gain (pp)} & \\
\midrule
\multicolumn{3}{@{}l}{\emph{Reported gains, matched comparisons}} \\
LVAgent~\cite{lvagent}               & $+3.05$  & \gbar{4.6} \\
VITAL~\cite{vital}                   & $+4.17$  & \gbar{6.3} \\
VideoSEAL~\cite{videoseal}           & $+5.83$  & \gbar{8.7} \\
Symphony~\cite{symphony}             & $+6.10$  & \gbar{9.2} \\
VideoAgent~\cite{videoagent}         & $+6.40$  & \gbar{9.6} \\
STAR~\cite{starbenchmarkreasoning}                     & $+6.40$  & \gbar{9.6} \\
LongVideoAgent~\cite{longvideoagent} & $+10.50$ & \gbar{15.8} \\
\addlinespace[2pt]
\textbf{Median} & $\mathbf{+6.10}$ & \\
\midrule
\multicolumn{3}{@{}l}{\emph{Re-execution variation, this work}} \\
Empirical execution spread $s_K^{(\mathrm{acc})}$ & 8.05 & \gbar{12.1} \\
Swing across $K_1,K_2$            & $11.38$ & \gbar{17.1} \\
\bottomrule
\end{tabular*}
\end{table}

% =====================================================================

\begin{table}[t]
\centering
\footnotesize
\setlength{\tabcolsep}{4.5pt}
\renewcommand{\arraystretch}{1.16}
\caption{\textbf{REPAIR separates directional reproducibility from
effect-size stability: two of five metrics reverse sign under
re-execution.}
Effects are Dynamic-SAGE minus Static-SAGE within an execution, so a
negative process effect means Dynamic-SAGE uses fewer steps or less time.
$\bar Z = \Delta_{K_1} - \Delta_{K_2}$ is the cross-execution shift in the
method effect; brackets give 95\% paired-bootstrap intervals on $\bar Z$.
$p_{\mathrm{Holm}}$ is Holm-adjusted across the five primary
cross-execution interaction tests. $^{\dagger}$ marks a per-execution
effect whose own interval covers zero. Every $\bar Z$ interval excludes
zero, so no metric is stable in magnitude across executions, and only
three preserve the sign of the effect.}
\label{tab:repair_claims}
\begin{tabular*}{\textwidth}{@{\extracolsep{\fill}}lrrrccl@{}}
\toprule
& \multicolumn{2}{c}{\textbf{Method effect}}
& \multicolumn{3}{c}{\textbf{Cross-execution shift}} & \\
\cmidrule(lr){2-3}\cmidrule(lr){4-6}
\textbf{Metric (unit)} &
$\boldsymbol{\Delta_{K_1}}$ &
$\boldsymbol{\Delta_{K_2}}$ &
$\boldsymbol{\bar Z}$ &
\textbf{95\% CI on} $\boldsymbol{\bar Z}$ &
$\boldsymbol{p_{\mathrm{Holm}}}$ &
\textbf{REPAIR assessment} \\
\midrule
\multicolumn{7}{@{}l}{\emph{Answer quality}} \\
Accuracy (pp)
& $+7.33$ & $-4.05$ & $+11.38$ & $[+8.03,+14.81]$ & $1.8{\times}10^{-10}$
& \rev{Direction reversal} \\
\addlinespace[2pt]
\multicolumn{7}{@{}l}{\emph{Process and cost}} \\
Reasoning turns
& $-0.920$ & $-0.680$ & $-0.241$ & $[-0.467,-0.010]$ & $3.6{\times}10^{-2}$
& Sign stable, magnitude shifts \\
Visible tool calls
& $-0.859$ & $-0.613$ & $-0.246$ & $[-0.444,-0.044]$ & $3.3{\times}10^{-2}$
& Sign stable, magnitude shifts \\
Primitive operations
& $+0.991$ & $+0.025^{\dagger}$ & $+0.966$ & $[+0.720,+1.224]$ & $2.2{\times}10^{-13}$
& Attenuates to null \\
Latency (s)
& $-3.492$ & $+21.314$ & $-24.806$ & $[-30.308,-19.981]$ & $1.0{\times}10^{-19}$
& \rev{Direction reversal} \\
\midrule
\multicolumn{6}{@{}l}{Metrics preserving effect sign across $K_1,K_2$}
& 3 of 5 \\
\bottomrule
\end{tabular*}
\end{table}

\subsection{A Significant Method Gain Reverses Under Re-Execution}
\label{sec:reexecution_accuracy}

Our central result is a reversal of the paired scientific conclusion
(Table~\ref{tab:reexecution_accuracy}). Dynamic-SAGE changes from a
significant $+7.33$-point advantage in $K_1$ to a significant
$-4.05$-point disadvantage in $K_2$, despite evaluation on the same
1,283 question--video pairs. The paired transitions likewise reverse: Dynamic-SAGE gains/loses
183/89 questions relative to Static-SAGE in $K_1$, versus 91/143 in
$K_2$. Thus, standard paired inference detects a difference within
each realized execution but does not establish that the difference
survives re-execution. REPAIR tests this directly. The cross-execution interaction is
$\bar Z=11.38$ percentage points, with 95\% paired-bootstrap CI
$[+8.03,+14.81]$ and
$t(1282)=6.597$, $p=6.11\times10^{-11}$. The two observed
execution-specific effects therefore differ significantly. The conclusion is unchanged when source video rather than question is
used as the resampling unit: the video-clustered 95\% CI is
$[+8.08,+14.85]$. The shift is broad: the Dynamic-minus-Static point estimate changes
from positive in $K_1$ to negative in $K_2$ across all 11 examined
subgroups spanning question format, modality, difficulty, and video
duration. Examples include long videos ($+9.80\rightarrow-6.76$),
visual questions ($+8.91\rightarrow-4.61$), and hard questions
($+9.39\rightarrow-3.64$); full subgroup results are provided in the
supplement. These results expose the distinction motivating REPAIR:
\emph{within-execution significance does not imply re-execution
stability}.

% =====================================================================
\subsection{Re-Execution Variation Exceeds Single-Run Uncertainty}
\label{sec:execution_variation}

The observed reversal is large relative to the uncertainty quantified
within either execution. The bootstrap standard errors of the
Dynamic-minus-Static accuracy effect are 1.27 and 1.18 percentage
points in $K_1$ and $K_2$, respectively, for a mean of 1.22 points.
By comparison, the empirical execution spread is
$s_K^{(\mathrm{acc})}=8.05$ points, approximately $6.6\times$ larger.
Because $K=2$, we use this spread descriptively rather than as a
precise estimate of a population execution-variance parameter. To contextualize this scale, we examine seven recent agentic VideoQA
systems for which the original paper reports a directly matched
comparison that adds or strengthens the agentic mechanism while
preserving the surrounding benchmark and evaluation setting
(Table~\ref{tab:recent_agentic_gains}). We use the absolute
within-paper accuracy difference between the matched baseline and
method; full provenance for each comparison is provided in the
supplementary material. The median matched gain is 6.10 points. Our 8.05-point empirical
execution spread is $1.32\times$ this median and exceeds six of the
seven reported gains, while the 11.38-point observed interaction is
$1.87\times$ the median and exceeds all seven. This comparison
contextualizes effect magnitude only; it does not imply that the
compared systems are themselves unstable.

\subsection{REPAIR Separates Reproducible Direction from Stable Effect Size}
\label{sec:repair_claims}

Applying REPAIR across the five primary outcomes reveals three
qualitatively different forms of re-execution behavior
(Table~\ref{tab:repair_claims}). Accuracy and latency exhibit
\emph{directional reversal}; reasoning turns and visible tool calls
retain their direction but exhibit significant \emph{magnitude
sensitivity}; and the primitive-operation effect \emph{attenuates}
from a significant increase to no measurable difference. All five cross-execution interactions remain significant after Holm
family-wise error correction (Table~\ref{tab:repair_claims}). Accuracy shows the strongest failure mode: a significant advantage in
$K_1$ becomes a significant disadvantage in $K_2$. Latency behaves
similarly. Dynamic-SAGE is 3.492\,s faster per question in $K_1$
(95\% CI $[-6.202,-0.785]$), but 21.314\,s slower in $K_2$
(95\% CI $[+17.496,+26.045]$), producing a
$-24.806$\,s interaction
(95\% CI $[-30.308,-19.981]$,
$p_{\mathrm{Holm}} = 1.02 \times 10^{-19}$). Reasoning turns and visible tool calls show a subtler pattern.
Dynamic-SAGE reduces both metrics in both executions, so their
directional claims reproduce. However, the effect sizes still change
significantly: reasoning turns shift from $-0.920$ to $-0.680$
($\bar Z=-0.241$, $p_{\mathrm{Holm}}=0.0364$), while visible calls shift from
$-0.859$ to $-0.613$ ($\bar Z=-0.246$, $p_{\mathrm{Holm}}=0.0329$). A repeated
evaluation can therefore preserve the qualitative conclusion while
rejecting effect-size invariance. Primitive operations exhibit attenuation rather than reversal. In
$K_1$, Dynamic-SAGE increases primitive operations by $+0.991$ per
question (95\% CI $[+0.784,+1.208]$); in $K_2$, the $+0.025$ effect
is indistinguishable from zero
(95\% CI $[-0.135,+0.184]$). The interaction remains large and highly
significant ($\bar Z=+0.966$, $p_{\mathrm{Holm}}=2.20\times10^{-13}$), showing that
an effect supported in one execution can effectively disappear in
another without changing nominal sign. These results illustrate why repeated evaluation should distinguish
\emph{directional reproducibility} from \emph{effect-size stability}.
A claim may reverse, preserve its direction while changing magnitude,
or attenuate to no measurable effect. The orchestration results also clarify what does reproduce.
Dynamic-SAGE reduces visible tool calls by 27.4\% in $K_1$ and 22.4\%
in $K_2$, but this compression does not translate into reproducible
resource savings. Primitive operations increase by 31.6\% in $K_1$
and remain approximately unchanged in $K_2$, while latency changes
from a 6.1\% reduction to a 64.4\% increase. The reproducible claim is
therefore that Dynamic-SAGE compresses its orchestration sequence, not
that it reproducibly reduces computation or runtime.

% =====================================================================
\subsection{Both Evaluation Arms Shift Under Re-Execution}
\label{sec:arm_reexecution}

The accuracy reversal could arise from Dynamic-SAGE varying against an
otherwise stable baseline. We therefore compare each system with itself
across executions (Table~\ref{tab:within_method_reexecution}).
Static-SAGE shifts $+4.75$ percentage points and Dynamic-SAGE
$-6.63$ points; 20.8\% and 23.5\% of their question-level correctness
labels change, respectively. Both shifts remain nonzero under
video-clustered resampling (Supplementary Sec.~S4). Their opposing movements compose the observed interaction:
$6.63+4.75=11.38$ percentage points. Thus, execution sensitivity is
present in both evaluation arms rather than arising from Dynamic-SAGE
varying against a fixed Static-SAGE reference.

\begin{table}[t]
\centering
\footnotesize
\setlength{\tabcolsep}{4pt}
\renewcommand{\arraystretch}{1.15}
\caption{\textbf{The two arms move in opposite directions under
re-execution, and item-level churn is roughly four times the net change.}
$\Delta_{\mathrm{re}}$ is the within-method accuracy change from $K_1$ to
$K_2$ on the same 1{,}283 items. Churn is the fraction of items whose
correctness label flips between executions; $p$ is an exact McNemar test
on the discordant items within each method. Net movement recovers only
$23\%$ and $28\%$ of the churn for Static- and Dynamic-SAGE respectively,
so aggregate accuracy conceals most of the instability. Because the two $\Delta_{\mathrm{re}}$ move in opposite directions, the magnitudes
of their shifts sum to the $11.38$~pp swing reported in
Tab.~\ref{tab:reexecution_accuracy}, which the reversal decomposes into
exactly.}
\label{tab:within_method_reexecution}
\begin{tabular*}{\linewidth}{@{\extracolsep{\fill}}lrrrrc@{}}
\toprule
& \multicolumn{2}{c}{\textbf{Accuracy (\%)}}
& \multicolumn{3}{c}{$\boldsymbol{K_1 \rightarrow K_2}$} \\
\cmidrule(lr){2-3}\cmidrule(l){4-6}
\textbf{Method} & $\boldsymbol{K_1}$ & $\boldsymbol{K_2}$
& $\boldsymbol{\Delta_{\mathrm{re}}}$ \textbf{(pp)}
& \textbf{Churn (\%)} & $\boldsymbol{p}$ \\
\midrule
Static-SAGE  & 60.80 & 65.55 & $+4.75$ & 20.81 & $2.4{\times}10^{-4}$ \\
Dynamic-SAGE & 68.12 & 61.50 & $-6.63$ & 23.46 & $1.3{\times}10^{-6}$ \\
\midrule
\multicolumn{3}{@{}l}{Combined swing (cf.\ Tab.~\ref{tab:reexecution_accuracy})}
& $11.38$ & & \\
\bottomrule
\end{tabular*}
\end{table}

% =====================================================================
\subsection{Limitations}
\label{sec:limitations}

Our audit uses $K=2$ executions of one system pair, orchestrator, and
benchmark. It therefore demonstrates that a significant paired
conclusion can reverse, but does not estimate the frequency or
population variance of such reversals; larger $K$ and broader systems
are required for that purpose. We also do not identify the source of
the observed movement, since provider-side serving state and mutable
external tools are outside our control. Accordingly, our results do
not establish whether Static- or Dynamic-SAGE is the better system.
Correctness also relies on automated semantic judging; although
judge-family robustness is reported in the supplement, we do not
provide human adjudication. Because unavailable source videos reduce the evaluable population,
our conclusions are restricted to the 1,283 matched pairs and do not
establish representativeness of the full SAGE-Bench population.
Finally, the literature comparison contextualizes effect magnitude
only and does not imply instability of the compared systems.

\section{Conclusion}
\label{sec:conclusion}

A statistically decisive paired result need not survive complete
re-execution. On the same 1,283 SAGE-Bench pairs, Dynamic-SAGE changes
from a significant $+7.33$-point advantage to a significant
$-4.05$-point disadvantage, with a significant 11.38-point
cross-execution interaction. REPAIR makes this missing evaluation dimension explicit by repeating
the paired comparison and testing whether its method effect changes.
Our audit reveals directional reversals, magnitude shifts, and effect
attenuation, while also showing that orchestration compression need
not imply computational savings. For execution-sensitive agentic
systems, question-level significance should therefore be accompanied
by evidence that the claimed effect survives re-execution.

%%%%%%%%%%%%%%%%%%%%%%%%%%%%%%%%%%%%%%%%%%%%%%%%%%%%%%%%%%%%%%%%%%%%%%%%%%%%%%%%
%% BACK MATTER
%%%%%%%%%%%%%%%%%%%%%%%%%%%%%%%%%%%%%%%%%%%%%%%%%%%%%%%%%%%%%%%%%%%%%%%%%%%%%%%%
\begin{kilrepro}
Both executions use the same \texttt{GPT-4o} orchestrator configuration,
prompts, reasoning budget, primitive and synthesized tool libraries, and
scoring protocol; the synthesized library is generated once and frozen
before either execution. All cross-execution analyses use the exact
intersection of 1{,}283 question--video pairs observed in both runs.
Full decoding settings appear in \cref{sec:supp_impl}, and the offline
synthesis procedure in \cref{sec:supp_synthesis}. The central finding is
that per-execution numbers are \emph{not} expected to reproduce exactly;
what should reproduce is a non-zero cross-execution interaction that is
large relative to the within-execution standard error.
\end{kilrepro}

\begin{contributions}
% TODO: replace with the actual division of labour before posting.
All authors contributed to the design of the evaluation protocol and to
the writing of the paper.
\end{contributions}

\begin{acknowledgments}
We thank the members of the Kurban Intelligence Lab for discussions and support.
\end{acknowledgments}

% \begin{funding}
% % TODO: add grant numbers and funders, or delete this block.
% \end{funding}

\begin{availability}
Code for the synthesis pipeline and the paired evaluation:
\ifbool{KIL@blind}
{\url{https://github.com/KurbanIntelligenceLab/Dynamic-SAGE}}.
SAGE-Bench is public and cited in \cref{sec:experiments}.
\end{availability}

\begin{conflicts}
The authors declare no competing interests.
\end{conflicts}

%%%%%%%%%%%%%%%%%%%%%%%%%%%%%%%%%%%%%%%%%%%%%%%%%%%%%%%%%%%%%%%%%%%%%%%%%%%%%%%%
\bibliography{references}

%%%%%%%%%%%%%%%%%%%%%%%%%%%%%%%%%%%%%%%%%%%%%%%%%%%%%%%%%%%%%%%%%%%%%%%%%%%%%%%%
\appendix
%%%%%%%%%%%%%%%%%%%%%%%%%%%%%%%%%%%%%%%%%%%%%%%%%%%%%%%%%%%%%%%%%%%%%%%%%%%%%%%%
%% APPENDIX -- migrated from the standalone supplementary.
%% Appendix sections are lettered automatically by the template; figures,
%% tables and equations restart at A1.
%%%%%%%%%%%%%%%%%%%%%%%%%%%%%%%%%%%%%%%%%%%%%%%%%%%%%%%%%%%%%%%%%%%%%%%%%%%%%%%%

This appendix supports the re-execution audit presented in the main text.
\cref{sec:supp_impl} reports the configuration held fixed across
executions. \cref{sec:supp_subgroups} reports the per-subgroup
Dynamic-minus-Static effects in both executions. \cref{sec:supp_judge}
describes the scoring protocol and judge-family robustness.
\cref{sec:supp_clustered} reports the video-clustered reanalysis.
\cref{sec:supp_synthesis} documents the offline synthesis procedure that
produced the audited tool library, and \cref{sec:supp_filtering} its
filtering statistics. \cref{sec:supp_dropped} documents the evaluation
population, the synthesis holdout, and the samples unavailable at each
execution. \cref{sec:supp_literature} gives the provenance of the matched
literature gains used in \cref{sec:execution_variation}.

Throughout, $K_1$ and $K_2$ denote the two complete executions of the
paired comparison, and $\hat{\Delta}_k$ denotes the Dynamic-minus-Static
effect within execution $K_k$.

\section{Implementation Details}
\label{sec:supp_impl}

This section lists the decoding and budget settings referenced in
\cref{sec:systems_configuration}. Both Static-SAGE and Dynamic-SAGE run the
\texttt{GPT-4o} orchestrator at temperature $0.0$ with a maximum
iterative reasoning budget of 10 steps. These settings are identical in
$K_1$ and $K_2$.

We emphasize that a nominal temperature of $0.0$ does not make a complete
agentic evaluation deterministic. Requests are served by a hosted model
whose serving configuration is not under our control, tool calls reach
external services whose responses may change between executions, and the
orchestrator's trajectory is path dependent: a single differing
intermediate observation changes the subsequent reasoning context and
can alter every later decision. The re-execution movement documented in
the main paper is consistent with this, and we make no claim to have
isolated which of these sources contributes most.

\paragraph{Configuration held fixed across executions.}
Across $K_1$ and $K_2$ we hold fixed the system and orchestration
prompts, decoding configuration, iterative reasoning budget, the
primitive library $\mathcal{T}_s$, the synthesized library
$\mathcal{T}_d$, the tool-registration procedure, and the scoring
protocol. The synthesized library is \emph{not} regenerated for $K_2$:
both executions register the identical five composite tools, so the
observed movement cannot be attributed to a different tool library.

\paragraph{What is not held fixed.}
Provider-side serving state, model-version routing behind the nominal
\texttt{GPT-4o} endpoint, and mutable external-tool state (in particular
web-search results, which change over time) are outside our control.
This is precisely the setting our audit targets: the reproducibility of
a complete realized evaluation under the configuration an author can
actually hold fixed.

\section{Per-Subgroup Effects in Both Executions}
\label{sec:supp_subgroups}

\cref{sec:reexecution_accuracy} states that the Dynamic-minus-Static point
estimate changes sign across all 11 examined subgroups.
Table~\ref{tab:subgroup_effects} reports these effects in full.

\begin{table}[t]
\centering
\footnotesize
\setlength{\tabcolsep}{4pt}
\renewcommand{\arraystretch}{1.12}
\caption{\textbf{Per-subgroup Dynamic-minus-Static accuracy effects
(pp) in both executions.} $N$ is the subgroup size within the 1,283
matched pairs. Every subgroup changes sign between $K_1$ and $K_2$.
Subgroup-level estimates are noisier than the aggregate and are reported
to show the breadth of the shift, not to support subgroup-specific
claims.}
\label{tab:subgroup_effects}
\begin{tabular}{@{}lrrrr@{}}
\toprule
\textbf{Subgroup} & \textbf{$N$} &
\textbf{$\hat{\Delta}_1$} & \textbf{$\hat{\Delta}_2$} &
\textbf{$\bar Z$} \\
\midrule
\multicolumn{5}{@{}l}{\textit{Question format}}\\
MCQ           & 586 & \posbadge{+5.29} & \negbadge{-3.24} & $+8.53$\\
Open-ended    & 697 & \posbadge{+9.04} & \negbadge{-4.73} & $+13.77$\\
\midrule
\multicolumn{5}{@{}l}{\textit{Modality}}\\
Visual        & 932 & \posbadge{+8.91} & \negbadge{-4.61} & $+13.52$\\
Verbal        &  75 & \posbadge{+2.67} & \negbadge{-1.33} & $+4.00$\\
Both          & 276 & \posbadge{+3.26} & \negbadge{-2.90} & $+6.16$\\
\midrule
\multicolumn{5}{@{}l}{\textit{Difficulty}}\\
Easy          & 385 & \posbadge{+4.68} & \negbadge{-3.90} & $+8.58$\\
Medium        & 322 & \posbadge{+6.72} & \negbadge{-4.60} & $+11.32$\\
Hard          & 526 & \posbadge{+9.39} & \negbadge{-3.64} & $+13.03$\\
Very hard     &  50 & \posbadge{+10.00} & \negbadge{-6.00} & $+16.00$\\
\midrule
\multicolumn{5}{@{}l}{\textit{Video duration}}\\
Short         & 681 & \posbadge{+5.14} & \negbadge{-1.62} & $+6.76$\\
Long          & 602 & \posbadge{+9.80} & \negbadge{-6.76} & $+16.56$\\
\bottomrule
\end{tabular}
\end{table}

The pattern to note is not the individual magnitudes, several of which
rest on small subgroups, but that no subgroup preserves the direction of
the $K_1$ conclusion. A subgroup analysis performed on $K_1$ alone would
have reported a consistent, broadly distributed Dynamic-SAGE advantage,
and the same analysis on $K_2$ alone would have reported a consistent
disadvantage. Neither localizes the effect to a question type, because
the movement is not question-type specific.

\section{Scoring Protocol and Judge Robustness}
\label{sec:supp_judge}

\paragraph{Scoring protocol.}
Final-answer correctness is scored by an automatic \texttt{GPT-4o}
semantic judge at temperature $0.0$, which compares each predicted
answer against the ground-truth answer. To prevent re-judging of an
identical response from itself introducing cross-execution variation,
the judge is applied once to the pooled set of all system--execution
outputs. Identical answer strings therefore receive identical
correctness labels wherever they occur. This does not make the semantic
judge infallible; rather, it ensures that an unchanged response cannot
receive a different label solely because it appeared in a different
execution.

\paragraph{Judge-family robustness.}
To check that the audited comparison is not an artifact of one judge
family, we re-score both executions with judges from three additional
model families (Table~\ref{tab:judge_validation}). Absolute accuracy
estimates vary across judges, as expected, but the finding of interest
is that the \emph{sign reversal between executions} is reproduced by
every judge: each ranks Dynamic-SAGE above Static-SAGE in $K_1$ and
below it in $K_2$. The reversal is therefore robust across the four
scorer families tested rather than being specific to the primary
\texttt{GPT-4o} judge.

\begin{table}[t]
\centering
\footnotesize
\setlength{\tabcolsep}{4pt}
\renewcommand{\arraystretch}{1.12}
\caption{\textbf{The reversal reproduces across judge families.}
Dynamic-minus-Static accuracy effect (pp) within each execution, scored
by four independent judges. \texttt{GPT-4o} is the primary evaluator
used throughout the paper. Absolute accuracies differ across judges, but
every judge reverses sign between $K_1$ and $K_2$.}
\label{tab:judge_validation}
\begin{tabular}{@{}lrrr@{}}
\toprule
\textbf{Judge} &
\textbf{$\hat{\Delta}_1$} &
\textbf{$\hat{\Delta}_2$} &
\textbf{$\bar Z$} \\
\midrule
\texttt{GPT-4o} (primary) & \posbadge{+7.33} & \negbadge{-4.05} & $+11.38$\\
\texttt{Claude Opus 4.6}  & \posbadge{+5.22} & \negbadge{-3.51} & $+8.73$\\
\texttt{Qwen3-235B}       & \posbadge{+5.83} & \negbadge{-3.977} & $+9.81$\\
\texttt{DeepSeek V4 Pro}  & \posbadge{+6.59} & \negbadge{-4.28} & $+10.87$\\
\bottomrule
\end{tabular}
\end{table}

We do not claim that automated semantic judging is equivalent to human
annotation. Validation against blinded human labels remains future work
and is stated as a limitation in \cref{sec:limitations}. The
fixed pooled scoring protocol prevents identical answer strings from
receiving different labels solely because they occur in different
executions, while the multi-judge analysis shows that the observed sign
reversal is robust across the scorer families tested.

\section{Video-Clustered Reanalysis}
\label{sec:supp_clustered}

SAGE-Bench draws multiple questions from some source videos, so
question-level resampling can understate uncertainty if outcomes are
correlated within a video. We therefore repeat the accuracy analyses
with a bootstrap that resamples source videos rather than individual
questions.

Table~\ref{tab:cluster_bootstrap} reports the resulting intervals.
The 1,283 matched questions correspond to 922 unique source videos,
with cluster sizes ranging from 1 to 5 questions (mean 1.39). For each
of 10,000 bootstrap iterations, we sample 922 source videos with
replacement and retain all questions associated with each sampled
video. If a video is sampled multiple times, its complete question set
is included multiple times. The four system--execution outcomes for
each question are resampled jointly, preserving their pairing.

\begin{table}[t]
\centering
\footnotesize
\setlength{\tabcolsep}{4pt}
\renewcommand{\arraystretch}{1.12}
\caption{\textbf{Video-clustered bootstrap robustness.}
The 1,283 matched questions correspond to 922 unique source videos.
Intervals use 10,000 cluster-bootstrap resamples of source videos.
All five intervals exclude zero.}
\label{tab:cluster_bootstrap}
\begin{tabular}{@{}lrr@{}}
\toprule
\textbf{Quantity} &
\textbf{Effect (pp)} &
\textbf{95\% CI} \\
\midrule
$K_1$: Dynamic$-$Static
& \posbadge{+7.33} & $[+4.89,+9.83]$ \\

$K_2$: Dynamic$-$Static
& \negbadge{-4.05} & $[-6.39,-1.73]$ \\

Cross-execution interaction
& $+11.38$ & $[+8.08,+14.85]$ \\

Static: $K_2-K_1$
& $+4.75$ & $[+2.31,+7.25]$ \\

Dynamic: $K_2-K_1$
& $-6.63$ & $[-9.38,-4.01]$ \\
\bottomrule
\end{tabular}
\end{table}

Clustering changes the intervals only slightly and leaves every
accuracy conclusion unchanged. The $K_1$ and $K_2$ effects remain
nonzero with opposite signs, the cross-execution interaction remains
nonzero, and both within-method re-execution shifts remain nonzero.

\section{Offline Synthesis Procedure}
\label{sec:supp_synthesis}

This section documents how the audited library $\mathcal{T}_d$ was
produced. It is included for completeness and reproducibility of the
audit target; the paper makes no claim that this procedure is superior
to alternative synthesis methods.

Algorithm~\ref{alg:offline_synthesis} gives the full procedure.
Synthesis is completed entirely offline, before either execution. The
orchestrator never writes Python at inference time, and the evaluation
runtime differs from Static-SAGE only by the additional registered tools
in $\mathcal{T}_d$. Offline synthesis processes the held-out set
$\mathcal{D}_h$ in batches of $15$ with a fixed random seed, and uses
agent-specific sampling temperatures: the Signature Agent uses
temperature $0.7$ to encourage diverse candidate interfaces, while the
Implementation Agent uses temperature $0.3$ for more stable executable
code. The verification stage is deterministic.

\paragraph{Implementation Agent prompt assembly.}
The Implementation Agent receives the existing tool sources, the
proposed docstring, the proposed signature, and the sibling signatures
from the same generation batch. Its output is restricted to imports and
an implementation body; the generator then assembles the complete
function using the signature and docstring produced by the Signature
Agent.

\paragraph{Verification Agent checks.}
The verifier first performs static checks, including Python syntax
validation, banned-import rejection, docstring validation, and broad
exception-handler rejection. It then imports the candidate in a
controlled test module and dry-runs the function using placeholder
inputs. To avoid expensive or side-effecting calls during verification,
the original SAGE tools are replaced with strict stubs that mimic their
runtime signatures and return shapes. This catches common implementation
errors such as invalid keyword arguments, non-dictionary returns,
missing docstrings, and passing transcript strings as media paths. The
verifier also enforces the \textit{evidence-return contract}, rejecting
tools that take a question-type argument or that invoke reasoning in
answer (verdict-committing) mode.

As noted in \cref{sec:limitations}, this establishes structural
validity under stubbed execution rather than functional correctness
under real execution; accepted tools are not separately validated
against ground-truth evidence on real videos.

\begin{algorithm}[tp]
\caption{Offline Dynamic Tool Synthesis}
\label{alg:offline_synthesis}
\small
\begin{algorithmic}[1]
\Statex \textbf{Input:} benchmark questions \(\mathcal{D}\), static tool library \(\mathcal{T}_{s}\), holdout size \(H\), batch size \(B\), max tools \(M\) (non-binding)
\Statex \textbf{Output:} dynamic tool library \(\mathcal{T}_{d}\)
\State \(\mathcal{D}_{h} \gets \textsc{StratifiedSample}(\mathcal{D}, H)\)
\State Exclude \(\mathcal{D}_{h}\) from final evaluation
\State \(\mathcal{P} \gets [\,]\)
\State \(\mathcal{T}_{d} \gets \emptyset\)
\State \(\mathcal{F} \gets \emptyset\)
\For{each batch \(Q_{b} \subset \mathcal{D}_{h}\) of size \(B\)}
    \State \(\mathcal{P}_{b} \gets \textsc{SignatureAgent}(Q_{b}, \mathcal{T}_{s}, \mathcal{P})\)
    \State Add name-unique proposals from \(\mathcal{P}_{b}\) to \(\mathcal{P}\)
\EndFor
\For{each proposal \(p \in \mathcal{P}\)}
    \If{\(|\mathcal{T}_{d}| = M\)}
        \State \textbf{break}
    \EndIf
    \State \(c \gets \textsc{ImplementWithRetries}(p, \mathcal{T}_{s}, \mathcal{P})\)
    \If{\(c\) fails verification} \Comment{syntax, safety, and evidence-return contract}
        \State Mark \(p\) as rejected
        \State \textbf{continue}
    \EndIf
    \State \(f \gets \textsc{Fingerprint}(c)\)
    \If{\(\textsc{PureWrapper}(f)\)}
        \State Mark \(p\) as skipped wrapper
    \ElsIf{\(f \in \mathcal{F}\)}
        \State Mark \(p\) as skipped duplicate
    \Else
        \State \(\mathcal{T}_{d} \gets \mathcal{T}_{d} \cup \{c\}\)
        \State \(\mathcal{F} \gets \mathcal{F} \cup \{f\}\)
    \EndIf
\EndFor
\State Register \(\mathcal{T}_{d}\) with the agent's tool library
\State \textbf{return} \(\mathcal{T}_{d}\)
\end{algorithmic}
\end{algorithm}

\paragraph{One-time synthesis overhead.}
For completeness we report the cost of producing $\mathcal{T}_d$, since
it is incurred once and is excluded from the per-question measurements
in the main text. Synthesis issued 72 LLM calls (10 Signature Agent, 62
Implementation Agent) consuming 384{,}156 prompt and 25{,}600 completion
tokens, at a total API cost of \$1.22. We do not present an amortization
or break-even analysis: such an analysis would presuppose a stable
per-question saving, and our central finding is that the
per-question resource effects are not reproducible across executions.

\section{Offline Synthesis Filtering}
\label{sec:supp_filtering}

Table~\ref{tab:synthesis_filtering} details the filtering pipeline.
Starting from 48 name-unique signature proposals, strict-stub validation
rejects 6 candidates. Of those passing validation, 37 are skipped as
duplicate composition fingerprints and none as pure wrappers, leaving 5
accepted composite tools registered in $\mathcal{T}_d$. The high
duplicate rate indicates that the proposal stage converges on a small
number of distinct recurring compositions.

\begin{table}[t]
\centering
\caption{Offline synthesis filtering. Each candidate is implemented and
validated first; the composition-fingerprint duplicate check and the
pure-wrapper check are applied only to candidates that pass validation.
From 48 name-unique signature proposals, the pipeline registers five
accepted composite tools.}
\label{tab:synthesis_filtering}
\begin{tabular}{lc}
\toprule
\textbf{Stage} & \textbf{Count} \\
\midrule
Raw signature proposals & 48 \\
Rejected by strict-stub validation & 6 \\
Skipped as duplicate composition fingerprint & 37 \\
Skipped as pure wrapper & 0 \\
Accepted into library & 5 \\
\bottomrule
\end{tabular}
\end{table}

\section{Evaluation Population}
\label{sec:supp_dropped}

SAGE-Bench contains 1,744 questions. We reserve 150 for offline tool
synthesis. Of the remaining 1,594, 290 had unavailable source videos at
the time of the original evaluation, leaving 1,304 questions in $K_1$.
At re-execution, 21 further source videos had become unavailable. All
cross-execution analyses in the main text therefore use the exact
intersection of \textbf{1,283 question--video pairs} observed in both
executions, so no execution effect can be attributed to a change in the
evaluation population.

Table~\ref{tab:sage_distribution} compares the 1,283-pair matched set
used for cross-execution inference with full SAGE-Bench. The matched
set does not skew appreciably toward any reported metadata category;
all category shares differ by at most approximately $0.33$ percentage
points from the full benchmark. This does not rule out selection
effects correlated with video availability rather than with these
metadata categories, so our conclusions are restricted to the 1,283
matched pairs rather than asserted to represent the full benchmark.

\begin{table}[t]
\centering
\small
\setlength{\tabcolsep}{6pt}
\caption{\textbf{Distribution comparison between full SAGE-Bench and
the 1,283-pair matched evaluation set.} The matched set excludes the
synthesis holdout and question--video pairs unavailable in either
execution.}
\label{tab:sage_distribution}
\begin{tabular}{lccccc}
\toprule
\textbf{Subset} & \textbf{SAGE N} & \textbf{SAGE \%} &
\textbf{Matched N} & \textbf{Matched \%} &
\textbf{$\Delta$ (pp)} \\
\midrule
MCQ & 793 & 45.47 & 586 & 45.67 & +0.20 \\
Open-ended & 951 & 54.53 & 697 & 54.33 & -0.20 \\
\midrule
Visual & 1266 & 72.59 & 932 & 72.64 & +0.05 \\
Verbal & 105 & 6.02 & 75 & 5.85 & -0.17 \\
Both & 373 & 21.39 & 276 & 21.51 & +0.12 \\
\midrule
Easy & 518 & 29.70 & 385 & 30.01 & +0.31 \\
Medium & 438 & 25.11 & 322 & 25.10 & -0.01 \\
Hard & 715 & 41.00 & 526 & 41.00 & 0.00 \\
Very hard & 73 & 4.19 & 50 & 3.90 & -0.29 \\
\midrule
Short videos & 920 & 52.75 & 681 & 53.08 & +0.33 \\
Long videos & 824 & 47.25 & 602 & 46.92 & -0.33 \\
\bottomrule
\end{tabular}
\end{table}

\paragraph{Synthesis holdout.}
Table~\ref{tab:holdout_distribution} reports the composition of the
150-question synthesis holdout relative to the matched evaluation set.
The holdout over-represents easy questions ($+8.7$ pp) and
under-represents hard ones ($-9.7$ pp). Because the holdout is used
only to propose reusable tool interfaces offline, while the audited
quantity is the cross-execution stability of a fixed library's paired
method effect, this imbalance cannot explain the change from $K_1$ to
$K_2$. We nevertheless report it for transparency.

\begin{table}[t]
\centering
\small
\setlength{\tabcolsep}{10pt}
\caption{Comparison of the 1,283-pair matched evaluation set and the
150-question holdout used for tool synthesis.}
\label{tab:holdout_distribution}
\begin{tabular}{lccc}
\toprule
\textbf{Subset} & \textbf{Matched (\%)} & \textbf{Holdout (\%)} & \textbf{$\Delta$ (pp)} \\
\midrule
MCQ & 45.7 & 48.0 & +2.3 \\
Open-ended & 54.3 & 52.0 & $-2.3$ \\
\midrule
Visual & 72.6 & 69.3 & $-3.3$ \\
Verbal & 5.8 & 9.3 & +3.5 \\
Both & 21.5 & 21.3 & $-0.2$ \\
\midrule
Easy & 30.0 & 38.7 & +8.7 \\
Medium & 25.1 & 27.3 & +2.2 \\
Hard & 41.0 & 31.3 & $-9.7$ \\
Very hard & 3.9 & 2.7 & $-1.2$ \\
\bottomrule
\end{tabular}
\end{table}

\paragraph{Video-level overlap.}
Because SAGE-Bench draws multiple questions from some source videos, the
question-level synthesis holdout does not guarantee video-level
disjointness: 122 of 146 distinct holdout videos also contribute
questions to the evaluation set. Synthesis is video-blind, operating
only on question text and the primitive tool signatures; it never
accesses video frames, transcripts, ground-truth answers, or evidence.
This overlap is nonetheless a limitation of the audited pipeline's
design, and we note it explicitly. It does not affect the re-execution
comparison, which holds the tool library fixed across both executions.

\section{Literature-Gain Provenance}
\label{sec:supp_literature}

\begin{table}[t]
\centering
\footnotesize
\setlength{\tabcolsep}{4pt}
\renewcommand{\arraystretch}{1.08}
\caption{\textbf{Provenance of matched literature gains used in
\cref{sec:execution_variation}.} Values are absolute within-paper accuracy improvements.
For multi-benchmark comparisons, benchmark-level differences are
averaged; LVAgent averages the two VideoMME subtitle conditions first. Here, M$n$ denotes the unweighted mean over $n$ benchmarks and
S denotes a single-benchmark comparison.}
\label{tab:literature_provenance}
\begin{tabular}{@{}lllllr@{}}
\toprule
System & Benchmark & Source & Baseline $\rightarrow$ Method & Agg. & Gain \\
\midrule
LVAgent & EgoSchema, LVB, MLVU, VMME
& T6 & Decide Agent $\rightarrow$ Dynamic Collab. & M4 & +3.05 \\

VITAL & VSI, Video-MMMU, MMVU
& T4 & w/o toolbox $\rightarrow$ VITAL-7B & M3 & +4.17 \\

VideoSEAL & MLVU, VMME, LVB, LVBench
& T1 & Coupled $\rightarrow$ Decoupled & M4 & +5.83 \\

Symphony & LVBench
& T4 & Planning $\rightarrow$ Full & S & +6.10 \\

VideoAgent & EgoSchema-500
& F3 & 1 round $\rightarrow$ 3 rounds & S & +6.40 \\

STAR & VideoMME, LongVideoBench
& T1--2 & GPT-4o (32f) $\rightarrow$ STAR & M2 & +6.40 \\

LongVideoAgent & LongTVQA+
& T2 & Agentic $\rightarrow$ AgenticRL & S & +10.50 \\
\bottomrule
\end{tabular}
\end{table}

\cref{sec:execution_variation} compares the scale of the observed
re-execution movement with matched gains reported by seven recent
agentic VideoQA systems. Table~\ref{tab:literature_provenance} gives
the provenance of each value: the benchmarks involved, the source
table or figure, and the specific baseline-to-method contrast used.
This comparison is intended only to contextualize effect-size
magnitude; it does not estimate the re-execution variability of those
systems.

When a matched comparison is reported on multiple benchmarks, we
compute the accuracy difference within each benchmark and report their
unweighted mean. For VideoMME entries with both subtitle and
no-subtitle conditions, the two conditions are first averaged within
VideoMME so that each benchmark contributes once.

The median of these seven matched gains is $6.10$ percentage points.
As emphasized in the main text, this comparison establishes only that
the movement observed in our audit is large relative to effect sizes
reported as meaningful improvements. It does not imply that any of the
compared systems would exhibit similar re-execution behavior.

\end{document}